# Video Highlights Detection and Summarization with Lag-Calibration based on Concept-Emotion Mapping of Crowd-sourced Time-Sync Comments

Qing Ping and Chaomei Chen
College of Computing & Informatics
Drexel University
{qp27, cc345}@drexel.edu

## Abstract

With the prevalence of video sharing, there are increasing demands for automatic video digestion such as highlight detection. Recently, platforms with crowdsourced time-sync video comments have emerged worldwide, providing a good opportunity for highlight detection. However, this task is non-trivial: (1) time-sync comments often lag behind their corresponding shot; (2) time-sync comments are semantically sparse and noisy; (3) to determine which shots are highlights is highly subjective. The present paper aims to tackle these challenges by proposing a framework that (1) uses concept-mapped lexical-chains for lag-calibration; (2) models video highlights based on comment intensity and combination of emotion and concept concentration of each shot; (3) summarize each detected highlight using improved SumBasic with emotion and concept mapping. Experiments on large real-world datasets show that our highlight detection method and summarization method both outperform other benchmarks with considerable margins.

## 1 Introduction

Every day, people watch billions of hours of videos on YouTube, with half of the views on mobile devices[1]. With the prevalence of video sharing, there is increasing demand for fast video digestion. Imagine a scenario where a user wants to quickly grasp a long video, without dragging the progress bar repeatedly to skip shots unappealing to the user. With automatically-generated highlights, users could digest the entire video in minutes, before deciding whether to watch the full video later. Moreover, automatic video highlight detection and summarization could benefit video indexing, video search and video recommendation.

However, finding highlights from a video is not a trivial task. First, what is considered to be a "highlight" can be very subjective. Second, a highlight may not always be captured by analyzing low-level features in image, audio and motions. Lack of abstract semantic information has become a bottleneck of highlight detection in traditional video processing.

Recently, crowdsourced time-sync video comments, or "bullet-screen comments" have emerged, where real-time generated comments will be flying over or besides the screen, synchronized with the video frame by frame. It has gained popularity worldwide, such as niconico in Japan, Bilibili and Acfun in China, YouTube Live and Twitch Live in USA. The popularity of the time-sync comments has suggested new opportunities for video highlight detection based on natural language processing.

Nevertheless, it is still a challenge to detect and label highlights using time-sync comments. First, there is almost inevitable lag for comments related to each shot. As in Figure 1, ongoing discussion about one shot may extend to next a few shots. Highlight detection and labeling without lag-calibration may cause inaccurate results. Second, time-sync comments are sparse semantically, both in number of comments per shot and number of tokens per comment. Traditionally bag-of-words statistical model may work poorly on such data.

---
[1] https://www.youtube.com/yt/press/statistics.html

Third, there is much uncertainty in highlight detection in an unsupervised setting without any prior knowledge. Characteristics of highlights must be explicitly defined, captured and modeled.

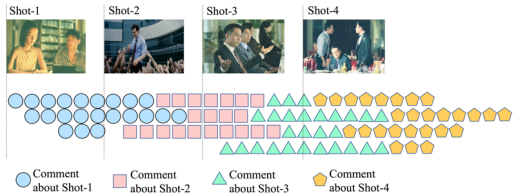

Figure 1.Lag Effect of Time-Sync Comments Shot by Shot.

To our best knowledge, little work has concentrated on highlight detection and labeling based on time-sync comments in unsupervised way. The most relevant work proposed to detect highlights based on topic concentration of semantic vectors of bullet-comments, and label each highlight with pre-trained classifier based on pre-defined tags (Lv, Xu, Chen, Liu, & Zheng, 2016). Nevertheless, we argue that emotion concentration is more important in highlight detection than general topic concentration. Another work proposed to extract highlights based on frame-by-frame similarity of emotion distribution (Xian, Li, Zhang, & Liao, 2015). However, neither work proposed to tackle the issue of lag-calibration, emotion-topic concentration balance and unsupervised highlight labeling simultaneously.

To solve these problems, the present study proposes the following: (1) word-to-concept and word-to-emotion mapping based on global word-embedding, from which lexical-chains are constructed for bullet-comments lag-calibration; (2) highlight detection based on emotional and conceptual concentration and intensity of lag-calibrated bullet-comments; (3) highlight summarization with modified Basic Sum algorithm that treats emotions and concepts as basic units in a bullet-comment.

The main contribution of the present paper are as follows: (1) We propose an entirely unsupervised framework for video highlight-detection and summarization based on time-sync comments; (2) We develop a lag-calibration technique based on concept-mapped lexical chains; (3) We construct large datasets for bullet-comment word-embedding, bullet-comment emotion lexicon and ground-truth for highlight-detection and labeling evaluation based on bullet-comments.

## 2 Related Work

### 2.1 Highlight detection by video processing

First, following the definition in previous work (M. Xu, Jin, Luo, & Duan, 2008), we define highlights as the most memorable shots in a video with high emotion intensity. Note that highlight detection is different from video summarization, which focuses on condensed storyline representation of a video, rather than extracting affective contents (K.-S. Lin, Lee, Yang, Lee, & Chen, 2013).

For highlight detection, some researchers propose to represent emotions in a video by a curve on the arousal-valence plane with low-level features such as motion, vocal effects, shot length, and audio pitch (Hanjalic & Xu, 2005), color (Ngo, Ma, & Zhang, 2005), mid-level features such as laughing and subtitles (M. Xu, Luo, Jin, & Park, 2009). Nevertheless, due to the semantic gap between low-level features and high-level semantics, accuracy of highlight detection based on video processing is limited (K.-S. Lin et al., 2013).

### 2.2 Temporal text summarization

The work in temporal text summarization is relevant to the present study, but also has differences. Some works formulate temporal text summarization as a constrained multi-objective optimization problem (Sipos, Swaminathan, Shivaswamy, & Joachims, 2012; Yan, Kong, et al., 2011; Yan, Wan, et al., 2011), as a graph optimization problem (C. Lin et al., 2012), as a supervised learning-to-rank problem (Tran, Niederée, Kanhabua, Gadiraju, & Anand, 2015), and as online clustering problem (Shou, Wang, Chen, & Chen, 2013).

The present study models the highlight detection as a simple two-objective optimization problem with constraints. However, the features chosen to evaluate the "highlightness" of a shot are different from the above studies. Because a highlight shot is observed to be correlated with high emotional intensity and topic concentration, coverage and non-redundancy are not goals of optimization any more, as in temporal text summarization. Instead, we focus on modeling emotional and topic concentration in present study.

### 2.3 Crowdsourced time-sync comment mining

Several works focused on tagging videos shot-by-shot with crowdsourced time-sync comments by manual labeling and supervised training (Ikeda, Kobayashi, Sakaji, & Masuyama, 2015), temporal and personalized topic modeling (Wu, Zhong, Tan, Horner, & Yang, 2014), or tagging video as a whole (Sakaji, Kohana, Kobayashi, & Sakai, 2016). One work proposes to generate summarization of each shot by data reconstruction jointly on textual and topic level (L. Xu & Zhang, 2017).

One work proposed a centroid-diffusion algorithm to detect highlights (Xian et al., 2015). Shots are represented by latent topics by LDA. Another work proposed to use pre-trained semantic vector of comments to cluster comments into topics, and find highlights based on topic concentration (Lv et al., 2016). Moreover, they use pre-defined labels to train a classifier for highlight labeling. The present study differs from these two studies in several aspects. First, before highlight detection, we perform lag-calibration to minimize inaccuracy due to comment lags. Second, we propose to represent each scene by the combination of topic and emotion concentration. Third, we perform both highlight detection and highlight labeling in unsupervised way.

### 2.4 Lexical chain

Lexical chains are a sequence of words in a cohesive relationship spanning in a range of sentences. Early work constructs lexical chains based on syntactic relations of words using the Roget's Thesaurus without word sense disambiguation (Morris & Hirst, 1991). Later work expands lexical chains by WordNet relations with word sense disambiguation (Barzilay & Elhadad, 1999; Hirst & St-Onge, 1998). Lexical chains is also constructed based on word-embedded relations for disambiguation of multi-words (Ehren, 2017). The present study constructs lexical chains for proper lag-calibration based on global word-embedding.

### 3 Problem Formulation

The problem in the present paper can be formulated as follows. The input is a set of time-sync comments, $C = \{c_1, c_2, c_3, \ldots, c_{|T|}\}$ with a set of timestamps $T = \{t_1, t_2, t_3, \ldots, t_{|T|}\}$ of a video $v$, a compression ratio $\tau_{highlight}$ for number of highlights to be generated, a compression ratio $\tau_{summary}$ for number of comments in each highlight summary. Our task is to (1) generate a set of highlight shots $S(v) = \{s_1, s_2, s_3, \ldots, s_n\}$, and (2) highlight summaries $A(v) = \{I_1, I_2, I_3, \ldots, I_n\}$ as close to ground truth as possible. Each highlight summary comprises a subset of all the comments in this shot: $I_i = \{c_1, c_2, c_3, \ldots, c_{n_i}\}$. Number of highlight shots $n$ and number of comments in summary $n_i$ are determined by $\tau_{highlight}$ and $\tau_{summary}$ respectively.

### 4 Video Highlight Detection

In this section, we introduce our framework for highlight detection. Two preliminary tasks are also described, namely construction of global time-sync comment word embedding and emotion lexicon.

### 4.1 Preliminaries

**Word-Embedding of Time-Sync Comments**

As pointed out earlier, one challenge in analyzing time-sync comments is the semantic sparseness, since number of comments and comment length are both very limited. Two semantically related words may not be related if they do not co-occur frequently in one video. To compensate, we construct a global word-embedding on a large collection of time-sync comments.

The word-embedding dictionary can be represented as: $D\{(w_1: v_1), (w_2: v_2), \ldots, (w_{|V|}: v_{|V|})$, where $w_i$ is a word, $v_i$ is the corresponding word-vector, $V$ is the vocabulary of the corpus.

**Emotion Lexicon Construction**

As emphasized earlier, it is crucial to extract emotions in time-sync comments for highlight detection. However, traditional emotion lexicons cannot be used here, since there exist too many Internet slangs that are specifically born on this type of platforms. For example, "23333" means "ha ha ha", and "6666" means "really awsome". Therefore, we construct an emotion lexicon tailored for time-sync comments from the word-embedding dictionary trained from last step. First we manually label words of the five basic emotional categories (*happy, anger, sad, fear and surprise*) as seeds (Ekman, 1992), from the top frequent words in the corpus. Here the sixth emotion category "*disgust*" is omitted because it is relatively rare in the dataset, and could be readily incorporated for other datasets. Then we expand the emotion lexicon by searching the top $N$ neighbors of each seed word in the word-embedding space, and adding a neighbor to seeds if the neighbor meets at least percentage of overlap $\gamma_{overlap}$ with all the seeds with minimum similarity of $sim_{min}$. The neighbors are searched based on cosine similarity in the word-embedding space.

## 4.2 Lag-Calibration

In this section, we introduce our method for lag-calibration following the steps of concept mapping, word-embedded lexical chain construction and lag-calibration.

**Concept Mapping**

To tackle the issue of semantic sparseness in time-sync comments, and to construct lexical-chains of semantically related words, words of similar meanings should be mapped to same concept first. Given a set of comments $C$ of video $v$, we first propose a mapping $\mathcal{F}$ from the vocabulary $V_C$ of comments $C$ to a set of concepts $K_C$, namely:

$$\mathcal{F}: V_C \to K_C \quad (|V_C| \geq |K_C|)$$

More specifically, mapping $\mathcal{F}$ maps each word $w_0$ into a concept $k = \mathcal{F}(w_0)$:

$$\mathcal{F}(w_0) = \mathcal{F}(w_1) = \mathcal{F}(w_2) = \cdots = \mathcal{F}(w_{|top\_n(w)|}) = \begin{cases} k, \exists k \in K_C \text{ and } \frac{|\{w|w \in top\_n(w_0) \wedge \mathcal{F}(w)=k\}|}{|top\_n(w_0)|} \geq \phi_{overlap} \\ w, \text{ otherwise} \end{cases} \quad (1)$$

and $top\_n(w_0)$ returns the top $n$ neighbors of word $w_0$ based on cosine similarity. For every word $w_0$ in comment $C$, we check percentage of its neighbors already mapped to a concept $k$. If the percentage exceeds the threshold $\phi_{overlap}$, then word $w_0$ together with its neighbors will be mapped to $k$. Otherwise they will be mapped to a new concept $w_0$.

**Lexical Chain Construction**

The next step is to construct all lexical chains in current time-sync comments of video $v$, so that lagged comments could be calibrated based on lexical chains. A lexical chain $l_{ij}$ comprises a set of triples $l_{ij} = \{(w, t, c)\}$, where $w$ is the actual mentioned word of concept $k_i$ in comment $c$, $t$ is the timestamp of the comment $c$. A lexical chain dictionary $D_{lexical\ chain}$ for time-sync comments $C$ of video $v$: $L_{lexical\ chain} = \{k_1: (l_{11}, l_{12}, l_{13} \ldots), k_2: (l_{21}, l_{22}, l_{23} \ldots), \ldots, k_{|K_C|}: (l_{|K_C|1}, l_{|K_C|2}, l_{|K_C|3} \ldots)\}$, where $k_i \in K_C$ is a concept, and $l_{ij}$ is the $j$th lexical chain of concept $k_i$. The algorithm for lexical chain construction is described in Algorithm 1.

Specifically, each comment in $C$ can be either appended to existing lexical chains, or added to new empty lexical chains, based on its temporal distance with existing chains controlled by Maximum silence $l_{max}$.

Note that word senses in the lexical chains constructed here are not disambiguated as most traditional algorithms do. Nevertheless, we argue that lexical chains are still useful, since our concept mapping is constructed from time-sync comments in its natural order, a progressively semantic continuity that naturally reinforces similar word senses for temporally close comments. This semantic continuity together with global word embedding ensures that our concept mapping is valid in most cases.

**Algorithm 1** Lexical Chain Construction

```
Input time-sync comments C. Word-to-concept
mapping F. Maximum silence l_max.
Output A dictionary of lexical chains
L_lexical chain.
Initialize L_lexical chain ← {}
for each c in C do
    t_current ← t_c
    for each word in c do
        k ← F(word)
        if k in L_lexical chain then
            chains ← L_lexical chain(k)
            t_previous ← t_chains[last]
            if t_current − t_previous ≤ l_max then
                chains[last] ← chains[last] ∪ c
            else
                chains ← chains ∪ {c}
            end if
        else
            L_lexical chain(k) ← {{c}}
        end if
    end for
end for
return L_lexical chain
```

Table 1. Lexical Chain Construction.

**Comment Lag-Calibration**

Now given constructed lexical chain dictionary $L_{lexical\ chain}$, we can calibrate the comments in $C$ based on their lexical chains. From our observation, the first comment about one shot usually occurs within the shot, while the rest may not be the case. Therefore, we calibrate the timestamp of each comment to the timestamp of first element of the lexical chain it belongs to. Among all the lexical chains (concepts) a comment belongs to, we pick the one with highest score $score_{k,c}$. $Score_{k,c}$ is computed as the sum frequency of each word in the chain weighted by its logarithm global frequency $\log(D(w).count)$. Therefore, each comment will be assigned to its most semantically important lexical-chain (concept) for calibration. The algorithm for the calibration is described in Algorithm 2.

Note that if there are multiple consecutive shots

**Algorithm 2** Lag-Calibration of Time-Sync Comments

```
Input time-sync comments C. Word-to-concept
mapping F. Lexical chain dictionary
L_{lexical chain}. Word-embedding dictionary D.
Output Lag-calibrated time-sync comments C'.
Initialize C' ← C
for each c in C' do
    chain_{best,c} ← {}
    score_{best,c} ← 0
    for each word in c do
        k ← F(word)
        chain_{k,c} ← L_{lexical chain}(k)[c]
        score_{k,c} ← 0
        for (w,t,c) in chain do
            N(w) ← D(w).count
            score_{k,c} ← score_{k,c} + 1/log(N(w))
        end for
        if score_{k,c} > score_{best} then
            chain_{best,c} ← chain_{k,c}
        end if
    end for
    t_c ← t_{chain_{best,c}[first]}
end for
return C'
```

Table 2. Lag-Calibration of Time-Sync Comments.

$\{s_1, s_2, ..., s_m\}$ with comments of similar contents, our lag-calibration method may calibrate many comments in shots $s_2, s_3, ..., s_m$ to the timestamp of the first shot $s_1$, if these comments are connected via lexical chains from shot $s_1$. This is not necessarily a bad thing since we hope to avoid selecting redundant consecutive highlight shots and leave opportunity for other candidate highlights, given a fixed compression ratio.

**Shot Importance Scoring**

In this section, we first segment comments by shots of equal temporal length $l_{scene}$, then we model shot importance. Then highlights could be detected based on shot importance.

A shot's importance is modeled to be impacted by two factors: comment concentration and commenting intensity. For comment concentration, as mentioned earlier, both concept and emotional concentration may contribute to highlight detection. For example, a group of concept-concentrated comments like "the background music/bgm/soundtrack of this shot is classic/inspiring/the best" may be an indicator of a highlight related to memorable background music. Meanwhile, comments such as "this plot is so funny/hilarious/lmao/lol/2333" may suggest a single-emotion concentrated highlight. Therefore, we combine these two concentrations in our model. First, we define emotional concentration $C_{emotion}$ of shot $s$ based on time-sync comments $C_s$ given emotional lexicon $E$ as follows:

$$C_{emotion}(C_s, s) = \frac{1}{-\sum_{e=1}^{5} p_e \cdot \log(p_e)} \quad (2)$$

$$p_e = \frac{|\{w|w \in C_s \wedge w \in E(e)\}|}{|C_s|} \quad (3)$$

Here we calculate the reverse of entropy of probabilities of five emotions within a shot as emotion concentration. Then we define topical concentration $C_{topic}$:

$$C_{topic}(C_s, s) = \frac{1}{-\sum_{k=1}^{|K_{C_s}|} p_k \cdot \log(p_k)} \quad (4)$$

$$p_k = \frac{\sum_{w \in C_s \wedge F(w)=k \wedge w \notin E} n_w / \log(N_w)}{\sum_{k \in K(C_s)} \sum_{w \in C_s \wedge F(w)=k \wedge w \notin E} n_w / \log(N_w)} \quad (5)$$

where we calculate the reverse of entropy of all concepts within a shot as topic concentration. The probability of each concept $k$ is determined by sum frequencies of its mentioned words weighted by their global frequencies, and divided by those values of all words in the shot.

Now the comment importance $\mathcal{I}_{comment}(C_s, s)$ of shot $s$ can be defined as:

$$\mathcal{I}_{comment}(s) = \lambda \cdot C_{emotion}(C_s, s) + (1 - \lambda) \cdot C_{topic}(C_s, s) \quad (6)$$

where $\lambda$ is a hyper-parameter, controlling the balance between emotion and concept concentration.

Finally, we define the overall importance of shot as:

$$\mathcal{I}(C_s, s) = \mathcal{I}_{comment}(C_s, s) \cdot \log(|C_s|) \quad (7)$$

Where $|C_s|$ is the length for all time-sync comments in shot $s$, which is a straightforward yet effective indicator of comment intensity per shot.

Now the problem of highlight detection can be modeled as a maximization problem:

$$Maximize \quad \sum_{s=1}^{N} \mathcal{I}(C_s, s) \cdot x_s \quad (8)$$

$$Subjective\ to \quad \begin{cases} \sum_{x=1}^{N} x_s \leq \tau_{highlight} \cdot N \\ x_s \in \{0,1\} \end{cases}$$

## 5  Video Highlight Summarization

Given a set of detected highlight shots $S(v) = \{s_1, s_2, s_3, ..., s_n\}$ of video $v$, each with all the lag-calibrated comments $C_s$ of that shot, we are at-tempting to generate summaries $A(v) = \{I_1, I_2, I_3, \ldots, I_n\}$ so that $I_s \subset C_s$ with compression ratio $\tau_{summary}$ and $I_s$ is as close to ground truth as possible.

We propose a simple but very effective summarization model, an improvement over SumBasic (Nenkova & Vanderwende, 2005) with emotion and concept mapping and two-level updating mechanism.

In the modified SumBasic, instead of only down-sampling the probabilities of words in a selected sentence to prevent redundancy, we down-sample the probabilities of both words and their mapped concepts for re-weighting each comment. This two-level updating mechanism could: (1) impose a penalty for sentences with semantically similar words to be selected; (2) still select a sentence with word already in the summary if this word occurs much more frequently. In addition, we use a parameter *emotion bias* $b_{emotion}$ to weight words and concepts when computing their probabilities, so that frequencies of emotional words and concepts will increase by $b_{emotion}$ compared to non-emotional words and concepts.

## 6 Experiment

In this section, we conduct experiments on large real datasets for highlight detection and summarization. We will describe the data collection process, evaluation metrics, benchmarks and experiment results.

### 6.1 Data

In this section, we describe the datasets collected and constructed in our experiments. All datasets and codes will be made publicly available on Github[2].

**Crowdsourced Time-sync Comment Corpus**

To train the word-embedding described in 4.1.1, we have collected a large corpus of time-sync comment from Bilibli[3], a content sharing website in China with time-sync comments. The corpus contains 2,108,746 comments, 15,179,132 tokens, 91,745 unique tokens, from 6,368 long videos. Each comment has 7.20 tokens on average.

Before training, each comment is first tokenized using Chinse word tokenization package Jieba[4]. Repeating characters in words such as "233333", "66666", "哈哈哈哈" are replaced with two same characters.

The word-embedding is trained using word2vec (Goldberg & Levy, 2014) with the skip-gram model. Number of embedding dimensions is 300, window size is 7, down-sampling rate is 1e-3, words with frequency lower than 3 times are discarded.

**Emotion Lexicon Construction**

After the word-embedding is trained, we manually select emotional words belonging to the five basic categories from the 500 most-frequent words in the word-embedding. Then we expand the emotion seeds iteratively using algorithm 1. After each expansion iteration, we also manually examine the expanded lexicon and remove inaccurate words to prevent the concept-drift effect, and use the filtered expanded seeds for expansion in next round. The minimum overlap $\gamma_{overlap}$ is set to be 0.05, and minimum similarity $sim_{min}$ is set to be 0.6. The selection of $\gamma_{overlap}$ and $sim_{min}$ is selected based on grid search in the range of [0,1]. The number of words for each emotion initially and after final expansion are listed in Table 3.

|       | Happy | Sad | Fear | Anger | Surprise |
|-------|-------|-----|------|-------|----------|
| Seeds | 17    | 13  | 21   | 14    | 19       |
| All   | 157   | 235 | 258  | 284   | 226      |

Table 3. Number of Initial and Expanded Emotion Words.

**Video Highlights Data**

To evaluate our highlight-detection algorithm, we have constructed a ground-truth dataset. Our ground-truth dataset takes advantage of user-uploaded mixed-clips about a specific video on Bilibli. Mixed-clips are a collage of video highlights by the user's own preferences. Then we take the most-voted highlights as ground-truth for a video.

The dataset contains 11 videos of 1333 minutes in length, with 75,653 time-sync comments in total. For each video, 3~4 video mix-clips about this video are collected from Bilibili. Shots that occur in at least 2 of all the mix-clips are considered as ground-truth highlights. All ground-truth highlights are mapped to the original video timeline, and the start and end time of the highlight are recorded as ground-truth. The mix-clips are selected based on the following heuristics: (1) The mixed-clips are searched on Bilibli using the keywords

---
[2] https://github.com/ChanningPing/VideoHighlightDetection
[3] https://www.bilibili.com/
[4] https://github.com/fxsjy/jieba

"video title + mixed clips"; (2) The mixed-clips are sorted by play times in descending order; (3) The mix-clip should be mainly about highlights of the video, not a plot-by-plot summary or gist; (4) The mix-clip should be under 10 minutes; (5) The mix-clip should contain a mix of several highlight shots instead of only one.

On average, each video has 24.3 highlight shots. The mean shot length of highlights is 27.79 seconds, while the mode is 8 and 10 seconds (frequency=19).

**Highlights Summarization Data**

We also construct a highlight-summarization (labeling) dataset of the 11 videos. For each highlight shot with its comments, we ask annotators to construct a summary of these comments by extracting as many comments as they see necessary. The rules of thumb are: (1) Comments of the same meaning will not be selected more than once; (2) The most representative comment for similar comments is selected; (3) If a comment stands out on its own, and is irrelevant to the current discussion, it will be discarded.

For 11 videos of 267 highlights, each highlight has on average 3.83 comments as its summary.

## 6.2 Evaluation Metrics

In this section, we introduce evaluation metrics for highlight-detection and summarization.

**Video Highlight Detection Evaluation**

For the evaluation of video highlight detection, we need to define what is a "hit" between a highlight candidate and reference. A rigid definition would be a perfect match of beginnings and ends between candidate and reference highlights. However, this is too harsh for any models. A more tolerant definition would be whether there is an overlap between a candidate and reference highlight. However, this will still underestimate model performance since users' selection of beginning and end of a highlight can be quite arbitrary some times. Instead, we propose a "hit" with relaxation $\varepsilon$ between a candidate $h$ and the reference $\hat{H}$ as follows:

$$hit_\varepsilon(h, \hat{H}) = \begin{cases} 1, & \exists \hat{h} \in \hat{H}: (s_h, e_h) \cap (s_{\hat{h}} - \varepsilon, e_{\hat{h}} + \varepsilon) \notin \emptyset \\ 0, & otherwise \end{cases} \quad (9)$$

Where $s_h, e_h$ is the start time and end time of highlight $h$, and $\varepsilon$ is the relaxation length of reference set $\hat{H}$. Further, the precision, recall and F-1 measure can be defined as:

$$Precision(H, \hat{H}) = \frac{\sum_{h=1}^{|H|} hit(h, \hat{H})}{|H|} \quad (10)$$

$$Recall(H, \hat{H}) = \frac{\sum_{\hat{h}=1}^{|\hat{H}|} hit(\hat{h}, H)}{|\hat{H}|} \quad (11)$$

$$F1(H, \hat{H}) = \frac{2 \cdot Precision(H, \hat{H}) \cdot Recall(H, \hat{H})}{Precision(H, \hat{H}) + Recall(H, \hat{H})} \quad (12)$$

In present study, we set the relaxation length to be 5 seconds. Also, the length for a candidate highlight is set to be 15 seconds.

**Video Highlight Summarization Evaluation**

We use ROUGE-1 and ROUGE-2 (C.-Y. Lin, 2004) as recall of candidate summary for evaluation:

$$ROUGE\text{-}n(C,R) = \frac{\sum_{\hat{S} \in R} \sum_{n\text{-gram} \in \hat{S}} Count_{match}(n\text{-gram})}{\sum_{\hat{S} \in R} \sum_{n\text{-gram} \in \hat{S}} Count(n\text{-gram})} \quad (13)$$

We use BLEU-1 and BLEU-2 (Papineni, Roukos, Ward, & Zhu, 2002) as precision. We choose BLEU for two reasons. First, a naïve precision metric will be biased for shorter comments, and BLEU can compensate this with the *BP* product factor:

$$BLEU\text{-}n(C, R) = BP \cdot \frac{\sum_{S \in C} \sum_{n\text{-gram} \in S} Count_{1V1-match}(n\text{-gram})}{\sum_{S \in C} \sum_{n\text{-gram} \in S} Count(n\text{-gram})} \quad (14)$$

$$BP = \begin{cases} 1, & if\ |C| > |R| \\ e^{(1-|R|/|C|)}, & if\ |C| \leq |R| \end{cases}$$

Where $C$ is the candidate summary and $R$ is the reference summary. Second, while reference summary contains no redundancy, candidate summary could falsely select multiple comments that are very similar and match to the same keywords in reference. In such case, the precision is extremely overestimated. BLEU will only count the match one-by-one, namely the number of match of a word will be the minimum frequencies in candidate and reference.

Finally, the F-1 measure can be defined as:

$$F1\text{-}n(C,R) = \frac{2 \cdot BLEU\text{-}n(C, R) \cdot ROUGE\text{-}n(C,R)}{BLEU\text{-}n(C, R) + ROUGE\text{-}n(C,R)} \quad (15)$$

## 6.3 Benchmark methods

**Benchmarks for Video Highlight Detection**

For highlight detection, we provide comparisons of different combinations of our model with three benchmarks:

- **Random-selection.** We select highlight shots randomly from all shots of a video.
- **Uniform-selection.** We select highlight shots at equal intervals.

- **Spike-selection**. We select those highlight shots who have the most number of comments within the shot.
- **Spike+E+T**. This is our method taking into consideration of emotion and topic concentration without the lag-calibration step.
- **Spike+L**. This is our method with only the lag-calibration step without taking into consideration of content concentration.
- **Spike+L+E+T**. This is our full model.

**Benchmarks for Video Highlight Summarization**

For highlight summarization, we provide comparisons of our method with five benchmarks:

- **SumBasic.** Summarization that exclusively exploits frequency for summary construction (Nenkova & Vanderwende, 2005).
- **Latent Semantic Analysis (LSA).** Summarization of text based on singular value decomposition (SVD) for latent topic discovery (Steinberger & Jezek, 2004).
- **LexRank**. Graph-based summarization that calculates sentence importance based on the concept of eigenvector centrality in a graph of sentences (Erkan & Radev, 2004).
- **KL-Divergence**. Summarization based on minimization of KL-divergence between summary and source corpus using greedy search (Haghighi & Vanderwende, 2009).
- **Luhn method**. Heuristic summarization that takes into consideration of both word frequency and sentence position in an article (Luhn, 1958).

### 6.4 Experiment Results

In this section, we report experimental results for highlight detection and highlight summarization.

**Results of Highlight Detection**

In our highlight detection model, the threshold for cutting a lexical chain $l_{max}$ is set to be 11 seconds, the threshold for concept mapping $\phi_{overlap}$ is set to be 0.5, threshold for concept mapping $top\_n$ is set to be 15, and the parameter $\lambda$ to control balance of emotion and concept concentration is set to be 0.9. A parameter analysis is provided in section 7.

The comparisons of precision, recall and F1 measures of different combinations of our method and the benchmarks are in Table 4. Our full model (Spike+L+E+T) outperforms all other benchmarks on all metrics. The precision and recall for Random-selection and uniform selection are low since they do not incorporate any structural or content information. Spike-selection improves considerably, since it takes advantage of the comment intensity of a shot. However, not all comment-intensive shots are highlights. For example, comments at the beginning and end of a video are usually high-volume greetings and goodbyes as a courtesy. Also, spike-selection usually condenses highlights on consecutive shots with high-volume comments, while our method could jump and scatter to other less intensive but emotionally or conceptually concentrated shots. This can be observed by the performance of Spike+E+T.

We also observe that lag-calibration (Spike+L) alone improves the performance of Spike-selection considerably, partially confirming our hypothesis that lag-calibration is important in time-sync comment related tasks.

|  | Precision | Recall | F-1 |
|---|---|---|---|
| Random-Selection | 0.1578 | 0.1587 | 0.1567 |
| Uniform-Selection | 0.1775 | 0.1830 | 0.1797 |
| Spike-Selection | 0.2594 | 0.2167 | 0.2321 |
| Spike+E+T | 0.2796 | 0.2357 | 0.2500 |
| Spike + L | **0.3125** | 0.2690 | 0.2829 |
| Spike+L+E+T | 0.3099 | **0.3071** | **0.3066** |

Table 4. Comparison of Highlight Detection Methods.

**Results of Highlight Summarization**

In our highlight summarization model, the emotional bias $b_{emotion}$ is set to be 0.3.

The comparisons on 1-gram BLEU, ROUGE and F1 of our method and the benchmarks are in Table 5. Our method outperforms all other methods, especially on ROUGE-1. LSA has lowest BLEU, mainly because LSA favors long and multi-word sentences statistically, however these sentences are not representative in time-sync com-ments. The SumBasic method also performs relatively poor since it considers semantically related words separately unlike our method that use concepts instead of words.

|  | **BLEU-1** | **ROUGE-1** | **F1-1** |
|---|---|---|---|
| LSA | 0.2382 | 0.4855 | 0.3196 |
| SumBasic | 0.2854 | 0.3898 | 0.3295 |
| KL-divergence | 0.3162 | 0.3848 | 0.3471 |
| Luhn | 0.2770 | 0.4970 | 0.3557 |
| LexRank | 0.3045 | 0.4325 | 0.3574 |
| Our method | **0.3333** | **0.6006** | **0.4287** |

Table 5. Comparison of Highlight Summarization Methods (1-Gram).

The comparisons on 2-gram BLUE, ROUGE and F1 of our method and the benchmarks are in Table 6. Our method also outperforms all other methods.

From the results, we believe that it is crucial to perform lag-calibration as well as concept and emotion mapping before summarization of time-sync comment texts. Lag-calibration shrinks prolonged comments to its original shots, preventing inaccurate highlight detection. Concept and emotional mapping works because time-sync comments are usually very short (7.2 tokens on average), the meaning of the comment is usually concentrated on one or two "central-words" in the comment. Emotion mapping and concept mapping could effectively prevent the redundancy in the generated summary.

|  | BLEU-2 | ROUGE-2 | F1-2 |
|---|---|---|---|
| SumBasic | 0.1059 | 0.1771 | 0.1325 |
| LSA | 0.0943 | 0.2915 | 0.1425 |
| LexRank | 0.1238 | 0.2351 | 0.1622 |
| KL-divergence | 0.1337 | 0.2362 | 0.1707 |
| Luhn | 0.1227 | 0.3176 | 0.1770 |
| Our method | **0.1508** | **0.3909** | **0.2176** |

Table 6. Comparison of Highlight Summarization Methods (2-Gram).

## 7 Influence of Parameters

### 7.1 Influence of Shot Length

We analyze the influence of *shot length* on $F1$ score for highlight detection. First from the distribution of highlight shot lengths in golden standards (Figure 2), we observe that most of the highlight shot lengths lie in the range of [0,25] (seconds), with 10 seconds as the mode. Therefore, we plot the $F1$ score of all four models at different shot lengths ranging from 5 to 23 seconds (Figure 3).

From Figure 3 we observe that (1) our method (Spike+L+E+T) consistently outperforms the other benchmarks at varied shot lengths; (2) however, the advantage of our method over Spike method seems to be moderated as the shot length increases. This is reasonable, because as the shot length becomes longer, the number of comments in each shot accumulates. After certain point, shot with significantly more comments will signify as highlight, no matter of the emotions and topics it contains. However, this may not always be the case. In reality, when there are too few comments, detection totally relying on volume will fail; on the other hand, when there are overwhelming volumes of comments evenly distributed among shots, spikes may not be a good indicator since every shot has equally large volumes of comments now. Moreover, most highlights in reality are below 15 seconds, and Figure 3 shows that our method could detect highlights more accurately at such finer level.

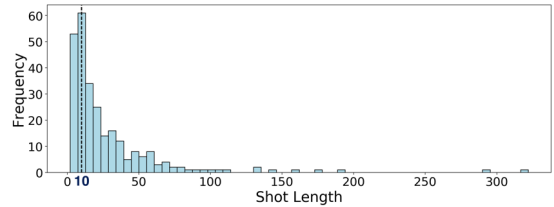

Figure 2. Distribution of Shot Lengths in Highlight Golden Standards.

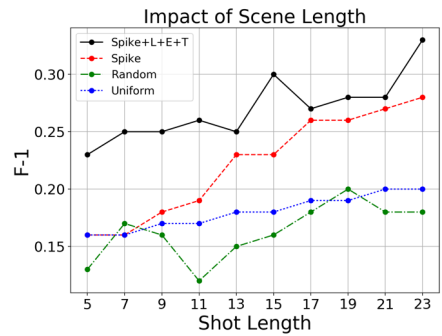

Figure 3. Influence of Shot Length on F-1 Scores of Highlight Detection.

### 7.2 Parameters for Highlight Detection

We analyze the influence of four parameters on recall for highlight detection: maximum silence for lexical chains $l_{max}$, the threshold for concept mapping $\phi_{overlap}$, the number of neighbors for concept mapping $top\_n$, and the balance of emotion and concept concentration $\lambda$ (Figure 4).

From Figure 4, we observe the following: (1) when it comes to lag-calibration, there seems to be an optimal *Max Silence Length*: 11 seconds as the longest blank continuance of a chain for our dataset. This value controls the compactness of a lexical chain. (2) In concept mapping, the *Minimum Overlap with Existing Concepts* controls the threshold for concept-merge, the higher the threshold the more similar the two merged concepts are. The recall increases as overlap increase to a certain point (0.5 in our dataset), and will not improve further after such point. (3) In concept mapping, there seems to be an optimal *Number of Neighbors* for searching (15 in our dataset). (4) The balance between emotion and concept concentration (*lambda*) is more on the emotion side (0.9 in our dataset).

### 7.3 Parameter for Highlight Summarization

We also analyze the influence of *emotion bias* $b_{emotion}$ on ROGUE-1 and ROGUE-2 for highlight summarization. The results are depicted in Figure 5.

From Figure 5, we observe that when it comes to highlight summarization, emotion plays a moderate role (emotion bias = 0.3). This is less significant than its role in the highlight detection task, where emotion concentration is much more important than concept concentration.

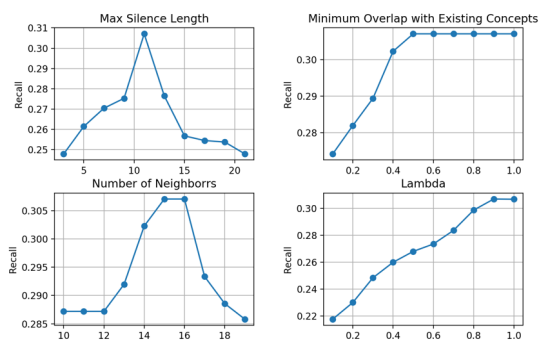

Figure 4. Influence of Parameters for Highlight Detection.

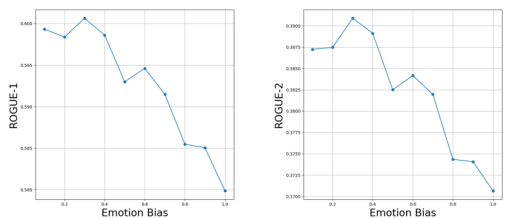

Figure 5. Influence of Parameter for Highlight Summarization.

## 8 Conclusion

In this paper, we propose a novel unsupervised framework for video highlight detection and summarization based on crowdsourced time-sync comments. For highlight detection, we develop a lag-calibration technique that shrinks lagged comments back to their original scenes based on concept-mapped lexical-chains. Moreover, video highlights are detected by scoring of comment intensity and concept-emotion concentration in each shot. For highlight summarization, we propose a two-level SumBasic that updates word and concept probabilities at the same time in each iterative sentence selection. In the future, we plan to integrate multiple sources of information for highlight detection, such as video meta-data, audience profiles, as well as low-level features of multiple modalities through video-processing.

## References

Barzilay, R., & Elhadad, M. (1999). Using lexical chains for text summarization. Advances in automatic text summarization, 111-121.

Ehren, R. (2017). Literal or idiomatic? Identifying the reading of single occurrences of German multiword expressions using word embeddings. Proceedings of the Student Research Workshop at the 15th Conference of the European Chapter of the Association for Computational Linguistics, 103–112.

Ekman, P. (1992). An argument for basic emotions. Cognition & emotion, 6(3-4), 169-200.

Erkan, G., & Radev, D. R. (2004). Lexrank: Graph-based lexical centrality as salience in text summarization. Journal of Artificial Intelligence Research, 22, 457-479.

Goldberg, Y., & Levy, O. (2014). word2vec explained: Deriving mikolov et al.'s negative-sampling word-embedding method. arXiv preprint arXiv:1402.3722.

Haghighi, A., & Vanderwende, L. (2009). Exploring content models for multi-document summarization. Paper presented at the Proceedings of Human Language Technologies: The 2009 Annual Conference of the North American Chapter of the Association for Computational Linguistics.

Hanjalic, A., & Xu, L.-Q. (2005). Affective video content representation and modeling. IEEE transactions on multimedia, 7(1), 143-154.

Hirst, G., & St-Onge, D. (1998). Lexical chains as representations of context for the detection and correction of malapropisms. WordNet: An electronic lexical database, 305, 305-332.

Ikeda, A., Kobayashi, A., Sakaji, H., & Masuyama, S. (2015). Classification of comments on nico nico douga for annotation based on referred contents. Paper presented at the Network-Based Information Systems (NBiS), 2015 18th International Conference on.

Lin, C., Lin, C., Li, J., Wang, D., Chen, Y., & Li, T. (2012). Generating event storylines from microblogs. Paper presented at the Proceedings of the 21st ACM international conference on Information and knowledge management.

Lin, C.-Y. (2004). Rouge: A package for automatic evaluation of summaries. Paper presented at the Text summarization branches out: Proceedings of the ACL-04 workshop.

Lin, K.-S., Lee, A., Yang, Y.-H., Lee, C.-T., & Chen, H. H. (2013). Automatic highlights extraction for drama video using music emotion and human face features. Neurocomputing, 119, 111-117.

Luhn, H. P. (1958). The automatic creation of literature abstracts. IBM Journal of research and development, 2(2), 159-165.

Lv, G., Xu, T., Chen, E., Liu, Q., & Zheng, Y. (2016). Reading the Videos: Temporal Labeling for Crowdsourced Time-Sync Videos Based on Semantic Embedding. Paper presented at the AAAI.

Morris, J., & Hirst, G. (1991). Lexical cohesion computed by thesaural relations as an indicator of the structure of text. Computational linguistics, 17(1), 21-48.

Nenkova, A., & Vanderwende, L. (2005). The impact of frequency on summarization. Microsoft Research, Redmond, Washington, Tech. Rep. MSR-TR-2005, 101.

Ngo, C.-W., Ma, Y.-F., & Zhang, H.-J. (2005). Video summarization and scene detection by graph modeling. IEEE Transactions on Circuits and Systems for Video Technology, 15(2), 296-305.

Papineni, K., Roukos, S., Ward, T., & Zhu, W.-J. (2002). BLEU: a method for automatic evaluation of machine translation. Paper presented at the Proceedings of the 40th annual meeting on association for computational linguistics.

Sakaji, H., Kohana, M., Kobayashi, A., & Sakai, H. (2016). Estimation of Tags via Comments on Nico Nico Douga. Paper presented at the Network-Based Information Systems (NBiS), 2016 19th International Conference on.

Shou, L., Wang, Z., Chen, K., & Chen, G. (2013). Sumblr: continuous summarization of evolving tweet streams. Paper presented at the Proceedings of the 36th international ACM SIGIR conference on Research and development in information retrieval.

Sipos, R., Swaminathan, A., Shivaswamy, P., & Joachims, T. (2012). Temporal corpus summarization using submodular word coverage. Paper presented at the Proceedings of the 21st ACM international conference on Information and knowledge management.

Steinberger, J., & Jezek, K. (2004). Using latent semantic analysis in text summarization and summary evaluation. Paper presented at the Proc. ISIM'04.

Tran, T. A., Niederée, C., Kanhabua, N., Gadiraju, U., & Anand, A. (2015). Balancing novelty and salience: Adaptive learning to rank entities for timeline summarization of high-impact events. Paper presented at the Proceedings of the 24th ACM International on Conference on Information and Knowledge Management.

Wu, B., Zhong, E., Tan, B., Horner, A., & Yang, Q. (2014). Crowdsourced time-sync video tagging using temporal and personalized topic modeling. Paper presented at the Proceedings of the 20th ACM SIGKDD international conference on Knowledge discovery and data mining.

Xian, Y., Li, J., Zhang, C., & Liao, Z. (2015). Video Highlight Shot Extraction with Time-Sync Comment. Paper presented at the Proceedings of the 7th International Workshop on Hot Topics in Planet-scale mObile computing and online Social neTworking.

Xu, L., & Zhang, C. (2017). Bridging Video Content and Comments: Synchronized Video Description with Temporal Summarization of Crowdsourced Time-Sync Comments. Paper presented at the Thirty-First AAAI Conference on Artificial Intelligence.

Xu, M., Jin, J. S., Luo, S., & Duan, L. (2008). Hierarchical movie affective content analysis based on arousal and valence features. Paper presented at the Proceedings of the 16th ACM international conference on Multimedia.

Xu, M., Luo, S., Jin, J. S., & Park, M. (2009). Affective content analysis by mid-level representation in multiple modalities. Paper presented at the Proceedings of the First International Conference on Internet Multimedia Computing and Service.

Yan, R., Kong, L., Huang, C., Wan, X., Li, X., & Zhang, Y. (2011). Timeline generation through evolutionary trans-temporal summarization. Paper presented at the Proceedings of the Conference on Empirical Methods in Natural Language Processing.

Yan, R., Wan, X., Otterbacher, J., Kong, L., Li, X., & Zhang, Y. (2011). Evolutionary timeline summarization: a balanced optimization framework via iterative substitution. Paper presented at the Proceedings of the 34th international ACM SIGIR conference on Research and development in Information Retrieval.